\documentclass[runningheads]{llncs}
\usepackage{graphicx}
\usepackage{cite}

\begin{document}
\title{Learning Shortcuts: On the Misleading Promise of NLU in Language Models}

\titlerunning{Misleading Promise of NLU}
%
\author{Geetanjali Bihani \inst{1}\and Julia Rayz \inst{1}}
\authorrunning{G. Bihani and Rayz J.}
\institute{Purdue University \\
\email{gbihani@purdue.edu, jtaylor1@purdue.edu}}
\maketitle              
\begin{abstract}
The advent of large language models (LLMs) has enabled significant performance gains in the field of natural language processing. However, recent studies have found that LLMs often resort to shortcuts when performing tasks, creating an illusion of enhanced performance while lacking generalizability in their decision rules. This phenomenon introduces challenges in accurately assessing natural language understanding in LLMs. Our paper provides a concise survey of relevant research in this area and puts forth a perspective on the implications of shortcut learning in the evaluation of language models, specifically for NLU tasks. This paper urges more research efforts to be put towards deepening our comprehension of shortcut learning, contributing to the development of more robust language models, and raising the standards of NLU evaluation in real-world scenarios.
\end{abstract}
\section{Introduction}

Large language models (LLMs) have become the convention in the field of natural language processing in recent years. The preference for LLMs can be attributed to their performance gains in a wide variety of NLP tasks, including question answering, textual entailment, sentiment analysis, and commonsense reasoning \cite{peters_deep_2018, devlin_bert_2019, sap_commonsense_2020}. As LLMs scale up, their performance gains not only compete with but also exceed human performance on language understanding benchmarks \cite{he2022debertav3, chowdhery2023palm}. Whether such performance gains are meaningful depends on the quality of the evaluation metrics and the relevance of benchmarking schemes \cite{marie2021scientific, bommasani2021opportunities}.

Unfortunately, LLMs have been shown to exploit spurious associations and dataset biases as `shortcuts', achieving inflated scores on NLU benchmarks, while lacking reliability and generalization on out-of-distribution samples. Recent works have shown that LLMs tend to learn shortcuts based on statistical cues e.g. the word ``not" \cite{niven_probing_2019}, keywords \cite{moon_masker_2021, du_towards_2021}  and cues related to language variations \cite{nguyen_learning_2021} to make predictions. This behavior, also known as \textit{shortcut learning}, leads the model to learn non-generalizable decision rules that do not perform well on out-of-distribution data \cite{moon_masker_2021, du_shortcut_2022}, but continues to give state-of-the-art results on independent identically distributed (IID) samples. Thus, PLM performance gains on NLU benchmarks do not necessarily portray improvement in the semantic and reasoning capabilities of language models.

This paper delineates the challenges in enhancing NLU capabilities within pre-trained language models amidst the presence of shortcut learning. We survey recent research on quantifying and mitigating shortcut learning and examine their implications for language under-standing in language models. Finally, we suggest potential research directions and guidelines that can help develop robust and reliable language models.

\section{What is Shortcut Learning?}\label{sec:2}
Shortcut learning refers to the phenomenon where models rely on superficial cues in the training data to make predictions instead of learning the underlying semantics to perform an NLU task. This over-reliance on specific features or biases results in poor generalization in out-of-distribution settings. Identifying shortcut learning in language models is an ongoing research area, with recent works utilizing model attention, dataset statistics, and human-annotated samples to identify spurious correlations \cite{moon_masker_2021, wang_identifying_2022}.

General-purpose neural language models have demonstrated learning incidental patterns present in natural language text, owing to the diverse linguistic cues embedded in their training corpora \cite{niven_probing_2019, nguyen_learning_2021}. While prior research asserted the robustness of LLMs in out-of-distribution detection and cross-domain generalization \cite{hendrycks_pretrained_2020}, recent studies have revealed that both LLMs and their fine-tuned versions rely on specific keyword-based shortcuts for classification \cite{moon_masker_2021}. Empirical analyses of LLMs highlight their dependence on spurious unigram and bigram cues to enhance Natural Language Inference (NLI) task performance \cite{niven_probing_2019}, as well as the utilization of syntactic heuristics for NLI tasks \cite{mccoy_right_2019}.

In an examination of fact verification classifiers, it was shown that claim-only models achieved comparable performance to evidence-aware models. This phenomenon was due to the former model's reliance on idiosyncrasies within the claims, which did not incorporate evidence in its predictions \cite{schuster_towards_2019}.

\subsection{Identifying Shortcuts}
Research dedicated to automatically identifying and mitigating spurious cues within training and fine-tuning data has been proposed in recent years \cite{wang_identifying_2020, tu_empirical_2020, wang_identifying_2022}. The understanding of shortcut learning behaviors in LLMs was recently extended by leveraging the long-tailed phenomenon \cite{du_towards_2021}. This work observed that LLMs tend to focus on the head of the word distribution, neglecting the poorly learned tail of the distribution. Furthermore, their findings indicate that shortcuts are predominantly learned in the early training iterations, impeding potential learning enhancements in later iterations.

While most works do not differentiate between spurious and genuine tokens, \cite{ren_huaslim_2023} addresses this distinction. The authors employ a human attention-guided approach to detect and alleviate shortcut learning. This mechanism identifies both model and dataset biases, utilizing human attention as a supervisory signal to compel the model to allocate more attention to 'relevant' tokens. Their approach enhances robustness on out-of-distribution data and maintains performance on IID data.

\section{Implications on NLU Evaluation}\label{sec:3}
\subsection{Examining the Impacts}
\subsubsection{Inflated NLU Performance Scores}
The performance gains of pre-trained language models on NLU tasks have been the subject of extensive scrutiny within the natural language processing domain. In several recent works, the performance of BERT-like models is shown to be entirely attributable to the exploitation of statistical cues. One work demonstrated that BERT uses the word "not" as a shortcut toward several NLU tasks, and the removal of such cues led to a significant drop in model performance, plummeting from 3 points below the human baseline to a random score \cite{niven_probing_2019}. Similarly, a drop of over $20\%$ in BERT-based NLU model performance was observed when evaluated on out-of-distribution datasets \cite{du_towards_2021}. Another study showed that if the majority of samples contain high levels of superficial information ($p>0.6$), PLM accuracy scores are inflated by $20\%$ \cite{he_unlearn_2019}.

Moreover, \cite{mccoy_right_2019} established that training and evaluating models using standard NLI datasets such as MNLI \cite{MNLI} result in models primarily relying on heuristics.  When assessed on a curated dataset lacking such heuristics, model performance dropped to $10\%$ for examples without heuristics, falling below chance performance. This highlights that NLI models may not inherently learn the correct rules for inference. Similarly, \cite{schuster_towards_2019} shows that even with a complete vocabulary overlap, the performance of fact verification models declined by $20\%$ when employing distinct language—devoid of shortcuts—while maintaining consistent claim and evidence relations (\textit{support} or \textit{refute}). \cite{ren_huaslim_2023} also revealed a substantial decline of $40\%-60\%$ in BERT-base model performance on NLI and fact verification tasks upon the removal of shortcuts. These studies underscore the pronounced dependence of LLMs on superficial information within the datasets. 

\subsubsection{Overconfidence in LLM Decisions}
Examining model performance extends beyond task accuracy metrics. Recent works also include a scrutiny of model confidence scores for incorrect predictions. \cite{ren_huaslim_2023} demonstrated that models tend to produce overconfident predictions for data samples with shortcut or trigger patterns, irrespective of the ground truth. This mismatch between model confidence and actual accuracy leads to the phenomenon known as miscalibration, impacting the reliability of models in real-world applications. Ideally, a well-calibrated model assigns high probabilities to correct predictions and low probabilities to incorrect decisions, aligning predicted probabilities with observed event frequencies.

The rising deployment of neural network architectures in high-risk real-world scenarios has prompted extensive research into their calibration \cite{thulasidasan_mixup_2019, malinin_predictive_2018, hendrycks_pretrained_2020}. Unfortunately, evaluations of neural network reliability indicate that their confidence predictions are often poorly calibrated and overly confident \cite{guo_calibration_2017, nixon2019measuring}.

Fine-tuning pre-trained language models exacerbates miscalibration \cite{kong_calibrated_2020, desai_calibration_2020, jiang_how_2021, bihani2023calibration}. This is attributed to the excessive parameterization of the models, leading to overfitting on the training data. The attention garnered by pre-trained language models stems from their inclination to exhibit increasing confidence during training, regardless of prediction accuracy \cite{chen_close_2022}. However, these models showcase calibration deterioration in out-of-domain scenarios \cite{desai_calibration_2020}. Notably, smaller models demonstrate improved calibration on in-domain data, while larger models exhibit better calibration on out-of-domain data \cite{dan_effects_2021}. These findings underscore the current inadequacies of pre-trained language models in terms of confidence calibration and reliability in decision-making.

\subsection{Improving NLU Amid Shortcut Learning}

The current focal paradigm in the area of improving natural language understanding and evaluation in LLMs revolves around alleviating shortcut learning to prevent inflated NLU task scores. Strategies such as ensemble learning and adversarial datasets are employed to diminish the impact of shortcuts, followed by the assessment of NLU task performance. More robust evaluations extend to measuring model generalization by assessing performance on out-of-distribution datasets.

\subsubsection{Data Centric Approaches}
Recent approaches to enhance model robustness and mitigate the impact of spurious cues involve the creation of datasets that contain curated examples that reduce model reliance on shortcuts during task learning. Incorporating adversarial data points during both training and evaluation, such as mirroring linguistic artifacts around labels has been shown to diminish the impact of spurious cues on model learning \cite{niven_probing_2019, schuster_towards_2019}. \cite{schuster_towards_2019} demonstrated a notable decrease in model performance when assessed on an adversarial test set, underscoring the necessity for Natural Language Understanding (NLU) evaluations to incorporate considerations for shortcut learning in LLMs. Additional efforts in this domain involve the development of adversarial datasets, such as HANS, which assesses whether NLI models rely on syntactic heuristics to accomplish the task \cite{mccoy_right_2019}.

More recently, novel data generation techniques have been proposed for mitigating spurious correlations and promoting generalization \cite{wu_generating_2022}. This technique involves training data generators to create and filter data points that exacerbate spurious correlations but yield significant performance drops in task accuracy. Moreover, even though data augmentation for bias removal reduces the model's dependence on spurious cues, it does not eliminate it \cite{zhou_towards_2020}. This persistence of lexical data biases emphasizes the ongoing challenges in effectively mitigating the impact of spurious cues in NLP models.

\subsubsection{Model Centric Approaches}
Various strategies centered on models to address shortcuts involve debiasing LLMs at the representation level \cite{karimi_mahabadi_end--end_2020}. This encompasses the creation of adversarial and Product-of-Expert (PoE) style ensembles, integrating both biased and robust models \cite{he_unlearn_2019, utama_mind_2020, stacey_there_2020}. In cases where prior knowledge about dataset biases is lacking, \cite{zhou_contrastive_2021} suggests that mitigating spurious cues is more effective when focusing on the model's predictive features. Thus, rather than debiasing input embeddings themselves, they achieve improved robustness by debiasing based on how the model utilizes embeddings.  In this scenario, the authors employ a simple Bag-of-Words (BoW) model to first capture superficial cues in the training data and then project the primary model embeddings into a space orthogonal to the BoW cue model. This orthogonality to superficial cue embeddings results in better debiasing. \cite{du_towards_2021} utilize a generalizable shortcut degree measurement, extracted from the dataset statistics, to smoothen softmax scores. This discourages the model from generating overly confident predictions for samples with higher shortcut degrees—indicative of the increased presence of shortcuts in a sample.

While these methods enhance performance on out-of-distribution samples, they have been shown to compromise the task accuracy on IID samples \cite{utama_mind_2020}. Furthermore, studies have highlighted that these methods inadvertently encode more biases into the inner representations of LMs \cite{karimi_mahabadi_end--end_2020}, creating newer shortcuts to deal with.

\section{Delving Deeper}\label{sec:4}
\subsection{Language and Vocabulary Variation}

Accurately identifying and eliminating specific shortcuts used by LLMs poses a significant challenge. This is exemplified by the reliance on the word "not" in BERT, where the removal of such shortcuts results in a notable drop in model performance. Successfully addressing this challenge requires nuanced strategies for pinpointing and mitigating these specific linguistic cues. Addressing the impact of shortcuts becomes complex when variations in language and vocabulary are introduced. Despite incorporating long-range context, encoding intricate lexical semantic phenomena for improved natural language understanding remains an ongoing endeavor \cite{fei2020retrofitting, bihani2021low, lexfit}. Even when the vocabulary overlap is significant, maintaining consistent performance poses a challenge. This raises questions regarding the broader applicability of current state-of-the-art LLMs across different language contexts and the necessity for models to exhibit resilience and adaptability in the face of varied linguistic expressions. 

\noindent{\textbf{A way forward -}} It is paramount to design robust methods capable of identifying and mitigating a diverse range of shortcut cues, both known and unknown.  A pivotal focus should be on enhancing the adaptability of state-of-the-art language models in diverse language contexts. This involves addressing tasks associated with language and vocabulary variations while categorizing different types of shortcuts across various contexts and task definitions. The emphasis must extend to mitigating challenges linked to unknown dataset biases and instances of incomplete task-specific knowledge during model training. Furthermore, these strategies need extension to navigate effectively through scenarios marked by limited information or unknown biases. This comprehensive approach is essential for a nuanced and effective mitigation of biases in model development.

\subsection{Out-of-Distribution Generalization}
Mitigating the impact of shortcuts becomes particularly challenging when LLMs are assessed on out-of-distribution datasets. The reliance on heuristics in NLI models, particularly when trained and evaluated on standard NLI datasets like MNLI \cite{MNLI}, poses a challenge to achieving robust inference capabilities. This is highlighted by a significant drop in performance when assessed on curated datasets lacking such heuristics, portraying difficulty in achieving generalization beyond the training distribution. Strategies and datasets need to be devised for the existing variety of NLU tasks, to enhance model robustness and adaptability to diverse data distributions.

\noindent{\textbf{A way forward -}} To enhance the overall generalization of language models on out-of-distribution datasets, it is crucial to formulate effective strategies and curate datasets specifically designed for this purpose. This necessitates a focused effort on defining specific robust inference capabilities, facilitating generalization beyond the confines of the training distribution. Furthermore, an in-depth exploration of methodologies that quantify the impact of shortcut removal on LLM performance is essential. Given the intricacies of datasets containing substantial amounts of superficial information, there is a need to develop nuanced approaches for comprehending the consequences of shortcut mitigation strategies. This involves a careful examination of trade-offs to ensure that performance does not experience significant deterioration. The aim is to strike a balance between addressing heuristics reliance and maintaining optimal model performance in the face of diverse and challenging datasets.

\subsection{Impact of Shortcut Removal}
Understanding the magnitude of the impact of shortcut removal on LLM performance is a crucial challenge. Dealing with datasets dominated by high levels of superficial information poses a significant challenge in mitigating shortcut learning impacts. The substantial decline of LLMs in the absence of a handful of such shortcuts underscores the need for a nuanced understanding of the consequences of shortcut mitigation strategies.

Du et al. \cite{du_towards_2021} also showed that model training exacerbates learning of superficial cues due to incentivizing reduction of training loss, which rapidly reduces when the model focuses on the head of word distribution, containing many shortcut features instead of the tail. This is reiterated in \cite{he_unlearn_2019} i.e. since the majority of training examples are not representative of the real-world data distribution (including the challenge data), minimizing the average training loss is not a legitimate objective. They solve this problem by modifying loss for known dataset biases but maintain that it is a challenge for unknown dataset biases and cases with incomplete task-specific knowledge.

\noindent{\textbf{A way forward -}}More works need to contribute towards measuring the impact of shortcut removal on LLM performance on NLU tasks. While most prior works utilize out-of-distribution accuracy, the types of datasets and the hardness of samples need to be systematized and defined to understand how different distribution shifts impact model learning behaviors.  

\subsection{The Caveats of Incentivizing Training Loss}
Understanding the impact of model training on the learning of superficial cues is crucial in the realm of natural language processing. Recent research demonstrated that model training exacerbates the learning of superficial cues \cite{du_towards_2021, he_unlearn_2019} because the optimization process incentivizes the reduction of training loss, which rapidly decreases when the model prioritizes the head of the word distribution, rich in shortcut features. Consequently, minimizing the average training loss may not be a legitimate objective. To address this issue, prior works propose modifying the loss function for known dataset biases. However, dealing with unknown dataset biases and cases with incomplete task-specific knowledge remains a challenge.

\noindent{\textbf{A way forward -}} There is an imperative need for the reevaluation of the incentives linked to training loss reduction in LLMs. We must delve into alternative training and post-processing strategies that prioritize the acquisition of meaningful linguistic patterns, rather than incentivizing the incorporation of shortcut features. This involves a fundamental shift in training objectives to better align with the actual distribution of real-world data. To ensure the reliability and accuracy of LLMs across diverse language tasks, it is crucial to investigate the interplay between model generalization and calibration. This exploration will contribute to the development of language models that generalize well and exhibit robustness against challenging samples within datasets.

\section{Conclusion}\label{sec:5}

This paper underscores the interplay between natural language understanding evaluation and shortcut learning within language models. We shed light on the distortions introduced by shortcut learning on NLU evaluation and emphasize the need for a nuanced understanding of its impacts. It is necessary to actively address the challenges posed by shortcut learning, both in terms of identifying and mitigating shortcuts and refining evaluation methodologies. Neglecting these issues jeopardizes the reliability and fairness of NLU assessments. As we move forward, collaborative efforts within the NLP community are essential to advance our comprehension of shortcut learning, fostering the development of more robust language models and elevating the standards of NLU evaluation in real-world contexts.

%
%
%
\bibliographystyle{splncs04}
\bibliography{sl_nlu}

%





\end{document}